\documentclass[11pt]{article}
\usepackage{eacl2017}
\usepackage{times}
\usepackage{url}
\usepackage{latexsym}
\usepackage{epsfig}

\usepackage{amsmath}
\usepackage{amssymb}

\usepackage[linesnumbered,vlined,ruled]{algorithm2e}

\makeatletter
\newcommand\footnoteref[1]{\protected@xdef\@thefnmark{\ref{#1}}\@footnotemark}
\makeatother

\eaclfinalcopy 
\def\eaclpaperid{107} 

\title{ShotgunWSD: An unsupervised algorithm for global word sense disambiguation inspired by DNA sequencing} 

\author{Andrei M. Butnaru, Radu Tudor Ionescu \and Florentina Hristea\\
  \\
  University of Bucharest\\
  Department of Computer Science\\
  14 Academiei, Bucharest, Romania\\
  {\tt butnaruandreimadalin@gmail.com}\\   
  {\tt raducu.ionescu@gmail.com}\\
  {\tt fhristea@fmi.unibuc.ro}  
}

\date{}

\begin{document}
\maketitle
\begin{abstract}
In this paper, we present a novel unsupervised algorithm for word sense disambiguation (WSD) at the document level. Our algorithm is inspired by a widely-used approach in the field of genetics for whole genome sequencing, known as the Shotgun sequencing technique. The proposed WSD algorithm is based on three main steps. First, a brute-force WSD algorithm 
is applied to short context windows (up to $10$ words) selected from the document in order to generate a short list of likely sense configurations for each window. In the second step, these local sense configurations are assembled into longer composite configurations based on suffix and prefix matching. The resulted configurations are ranked by their length, and the sense of each word is chosen based on a voting scheme that considers only the top $k$ configurations in which the word appears. We compare our algorithm with other state-of-the-art unsupervised WSD algorithms and demonstrate better performance, sometimes by a very large margin. We also show that our algorithm can yield better performance than the Most Common Sense (MCS) baseline on one data set. Moreover, our algorithm has a very small number of parameters, is robust to parameter tuning, and, unlike other bio-inspired methods, it gives a deterministic solution (it does not involve random choices). 
\end{abstract}

\section{Introduction}

Word Sense Disambiguation (WSD), the task of identifying which sense of a word is used in a given context, is a core NLP problem, having the potential to improve many applications such as machine translation~\cite{Carpuat-EMNLP-2007}, text summarization~\cite{Plaza-WSD-2011}, information retrieval~\cite{adi-radu-WSD-2012,Chifu-IPM-2014} or sentiment analysis~\cite{Sumanth-WASSA-2015}. Most of the existing WSD algorithms~\cite{agirre-2006,Navigli-WSD-2009} are commonly classified into supervised, unsupervised, and knowledge-based techniques, but hybrid approaches have also been proposed in the literature~\cite{Hristea-2008}. The main disadvantage of supervised methods (that have led to the best disambiguation results) is that they require a large amount of annotated data which is difficult to obtain. Hence, over the last few years, many researchers have concentrated on developping unsupervised learning approaches~\cite{Schwab-COLING-2012,Schwab-WET-2013,GETALP-2013,Chen-EMNLP-2014,Bhingardive-NAACL-2015}. In this paper, we introduce a novel WSD algorithm, termed ShotgunWSD\footnote{Our open source Java implementation of ShotgunWSD is freely available at http://ai.fmi.unibuc.ro/Home/Software.}, that stems from the Shotgun genome sequencing technique~\cite{Shotgun-1981,Istrail-2004}. Our WSD algorithm is also unsupervised, but it requires knowledge in the form of WordNet~\cite{Miller-WN-1995,Fellbaum-WN-1998} synsets and relations as well. Thus, our algorithm can be regarded as a hybrid approach. 

WSD algorithms can perform WSD at the local or at the global level. A local WSD algorithm, such as the extended Lesk measure~\cite{Lesk-1986,Banerjee-CICLING-2002,Banerjee-IJCAI-2003}, is designed to assign the appropriate sense, from an existing sense inventory, for a target word in a given context window of a few words. For instance, for the word ``sense'' in the context ``You have a good sense of humor.'', the sense that corresponds to the \emph{natural ability} rather than the \emph{meaning of a word} or the \emph{sensation} should be chosen by a WSD algorithm. Rather more generally, a global WSD approach aims to choose the appropriate sense for each ambiguous word in a text document. The straightforward solution is the exhaustive evaluation of all sense combinations (configurations)~\cite{Patwardhan-CICLing-2003}, but the time complexity is exponential with respect to the number of words in the text, as also noted by \newcite{Schwab-COLING-2012}, \newcite{Schwab-WET-2013}. Indeed, the brute-force (BF) solution quickly becomes impractical for windows of more than a few words. Hence, several approximation methods have been proposed for the global WSD task in order to overcome the exponentional growth of the search space~\cite{Schwab-COLING-2012,Schwab-WET-2013}. Our algorithm is designed to perform global WSD by combining multiple local sense configurations that are obtained using BF search, thus avoiding BF search on the whole text. A local WSD algorithm is employed to build the local sense configurations. We alternatively use two methods at this step, namely the extended Lesk measure~\cite{Banerjee-CICLING-2002,Banerjee-IJCAI-2003} and an approach based on deriving sense embeddings from word embeddings~\cite{Bengio-JMLR-2003,Collobert-ICML-2008,Mikolov-NIPS-2013}. Both local WSD approaches are based on WordNet synsets and relations.

Our global WSD algorithm can be briefly described in a few steps. In the first step, context windows of a fixed length $n$ are selected from the document, and for each context window the top scoring sense configurations constructed by BF search are kept for the second step. The retained sense configurations are merged based on suffix and prefix matching. The configurations obtained so far are ranked by their length (the longer, the better), and the sense of each word is given by a majority vote on the top $k$ configurations that cover the respective word. Compared to other state-of-the-art bio-inspired methods~\cite{Schwab-COLING-2012,Schwab-WET-2013}, our algorithm has less parameters. Different from the other methods, these parameters ($n$ and $k$) can be intuitively tuned with respect to the WSD task. 
As we select a single context window at every possible location in a text, our algorithm becomes deterministic, obtaining the same global configuration for a given set of parameters and input document. Thus, our algorithm is not affected by random chance, unlike stochastic algorithms such as Ant Colony Optimization~\cite{Lafourcade-ACO-2010,Schwab-COLING-2012,Schwab-WET-2013}.

We have conducted experiments on SemEval 2007~\cite{SemEval-task7-2007}, Senseval-2~\cite{Senseval2-2001} and Senseval-3~\cite{Senseval3-2004} data sets in order to compare ShotgunWSD with three state-of-the-art approaches~\cite{Schwab-WET-2013,Chen-EMNLP-2014,Bhingardive-NAACL-2015} along with the Most Common Sense (MCS) baseline\footnote{Also known as the Most Frequent Sense baseline.}, which is considered as the strongest baseline in WSD~\cite{agirre-2006}. The empirical results show that our algorithm compares favorably to these approaches.

The rest of this paper is organized as follows. Related work on unsupervised WSD algorithms is presented in Section~\ref{sec_RelatedWork}. The ShotgunWSD algorithm is described in Section~\ref{sec_ShotgunWSD}. The experiments are given in Section~\ref{sec_Experiments}. Finally, we draw our conclusions in Section~\ref{sec_Conclusions}.

\section{Related Work}
\label{sec_RelatedWork}

There is a broad range of methods designed to perform WSD~\cite{agirre-2006,Navigli-WSD-2009,VidhuBhala-AIR-2014}. The most accurate techniques are supervised~\cite{Navigli-ACL-2016}, but they require annotated training data which is not always available. In order to overcome this limitation, some researchers have proposed various unsupervised or knowledgde-based WSD methods~\cite{Banerjee-CICLING-2002,Banerjee-IJCAI-2003,Schwab-COLING-2012,Nguyen-AIR-2013,Schwab-WET-2013,GETALP-2013,Chen-EMNLP-2014,Agirre-COLI-2014,Bhingardive-NAACL-2015}. Since our approach is unsupervised and based on WordNet~\cite{Miller-WN-1995,Fellbaum-WN-1998}, our focus is to present related work in the same direction. \newcite{Banerjee-CICLING-2002} extend the gloss overlap algorithm of \newcite{Lesk-1986} by using WordNet relations. \newcite{Patwardhan-CICLing-2003} proposed a brute-force algorithm for global WSD by employing the extended Lesk measure~\cite{Banerjee-CICLING-2002,Banerjee-IJCAI-2003} to compute the semantic relatedness among senses in a given text. However, their BF approach is not suitable for whole text documents, because of the high computational time. More recently, \newcite{Schwab-COLING-2012} have proposed and compared three stochastic algorithms for global WSD as alternatives to BF search, namely a Genetic Algorithm, Simulated Annealing, and Ant Colony Optimization. Among these, the authors~\cite{Schwab-COLING-2012,Schwab-WET-2013} have found that the Ant Colony Algorithm yields better results than the other two.

Recently, word embeddings have been used for WSD~\cite{Chen-EMNLP-2014,Bhingardive-NAACL-2015,Navigli-ACL-2016}. Word embeddings are well known in the NLP community~\cite{Bengio-JMLR-2003,Collobert-ICML-2008}, but they have recenlty become more popular due to the work of \newcite{Mikolov-NIPS-2013} which introduced the \emph{word2vec} framework that allows to efficiently build vector representations from words. \newcite{Chen-EMNLP-2014} present a unified model for joint word sense representation and disambiguation. They use the Skip-gram model to learn word vectors. On the other hand, \newcite{Bhingardive-NAACL-2015} use pre-trained word vectors to build sense embeddings by averaging the word vectors produced for each sense of a target word. As their goal is to find an approximation for the MCS baseline, they try to find the sense embedding that is closest to the embedding of the target word. \newcite{Navigli-ACL-2016} propose different methods through which word embeddings can be leveraged in a \emph{supervised} WSD system architecture. Remarkably, they find that a WSD approach based on word embeddings alone can provide significant performance improvements over a state-of-the-art WSD system that uses standard WSD features.

\section{ShotgunWSD}
\label{sec_ShotgunWSD} 

As also noted by \newcite{Schwab-COLING-2012}, brute-force WSD algorithms based on semantic relatedness~\cite{Patwardhan-CICLing-2003} are not practical for whole text disambiguation due to their exponential time complexity. In this section, we describe a novel WSD algorithm that aims to avoid this computational issue. 
Our algorithm is inspired by the Shotgun genome sequencing technique~\cite{Shotgun-1981} which is used in genetics research to obtain long DNA strands. For example, \newcite{Istrail-2004} have used this technique to assemble the human genome. Before a long DNA strand can be read, Shotgun sequencing needs to create multiple copies of the respective strand. Next, DNA is randomly broken down into many small segments called \emph{reads} (usually between $30$ and $400$ nucleotides long), by adding a restriction enzyme into the chemical solution containing the DNA. The reads can then be sequenced using Next-Generation Sequencing techonlogy~\cite{Karl-NGS-2009}, for example by using an Illumina (Solexa) machine~\cite{Solexa-2004}. In genome assembly, the low quality reads are usually eliminated~\cite{NGS-QC-PLoS-2012} and the whole genome is reconstructed by assembling the remaining reads. One strategy is to merge two or more reads in order to obtain longer DNA segments, if they have a significant overlap of matching nucleotides. Because of reading errors or mutations, the overlap is usually measured using a distance measure, e.g. edit distance~\cite{levenshtein-1966}. If a backbone DNA sequence is available, the reads are aligned to the backbone DNA before assembly, in order to find their approximate position in the DNA that needs to be reconstructed.

\begin{algorithm*}[!th]
\small{
\caption{ShotgunWSD Algorithm\label{alg_ShotgunWSD}}

\textbf{Input}: 

$D = (w_1, w_2, ..., w_m)$ -- a document of $m$ words denoted by $w_i, i \in \{1,2, ..., m\}$;

$n$ -- the length of the context windows ($1 < n < m$);

$k$ -- the number of sense configurations considered for the voting scheme ($k > 0$);

\BlankLine
\textbf{Initialization}:

$c \leftarrow 20$\;

\For{$i \in \{1,2,...,m\}$}
{		
		$S_{w_i} \leftarrow$ the set of WordNet synsets of $w_i$\;
}

$\mathcal{S} \leftarrow \emptyset$\;

$G \leftarrow (0,0,....,0)$, such that $|G| = m$\;

\BlankLine
\textbf{Computation}:

\For{$i \in \{1,2, ..., m-n+1\}$}
{
		$\mathcal{C}_i \leftarrow \emptyset$\;
		
		\While{did not generate all sense configurations}
		{
				$C \leftarrow$ a new configuration $(s_{w_i}, s_{w_{i+1}},..., s_{w_{i+n-1}}), s_{w_{j}} \in S_{w_{j}},$ $\forall j \in \{i,...,i+n-1\},$ such that $C \notin \mathcal{C}_i$\;
				
				$r \leftarrow 0$\;
				
				\For{$p \in \{1,2,...,n-1\}$}
				{
						\For{$q \in \{p+1,2,...,n\}$}
						{
								$r \leftarrow r + \mbox{relatedness}(C[p], C[q])$\;
						}
				}
				
				$\mathcal{C}_i \leftarrow \mathcal{C}_i \cup \{(C, i, n, r)\}$\;
		}
		
		$\mathcal{C}_i \leftarrow$ the top $c$ configurations obtained by sorting the configurations in $\mathcal{C}_i$ by their relatedness score (descending)\;
		
		$\mathcal{S} \leftarrow \mathcal{S} \cup \mathcal{C}_i $\;
}

\For{$l \in \{min\{4,n-1\},...,1\}$}
{
		\For{$p \in \{1,2,...,|\mathcal{S}|\}$}
		{
				$(C_p, i_p, n_p, r_p) \leftarrow$ the $p$-th component of $\mathcal{S}$\;
	
				\For{$q \in \{1,2,...,|\mathcal{S}|\}$}
				{
						$(C_q, i_q, n_q, r_q) \leftarrow$ the $q$-th component of $\mathcal{S}$\;
				
						\If{$i_q - i_p < n_p$ and $i_p \neq i_q$}
						{
								$t \leftarrow \mbox{true}$\;
								
								\For{$x \in \{1,...,l\}$}
								{
										\If{$C_p[n_p-l+x] \neq C_q[x]$}
										{
												$t \leftarrow \mbox{false}$\;
										}
								}
								
								\If{$t = true$}
								{
										$C_{p \oplus q} \leftarrow (C_p[1],C_p[2],..., C_p[n_p], C_q[l+1],C_q[l+2],...,C_q[n_q])$\;
									
										$r_{p \oplus q} \leftarrow r_p$\;
									
										\For{$i \in \{1,2,...,n_p + n_q - l\}$}
										{
												\For{$j \in \{l+1,l+2,...,n_q\}$}
												{
													$r_{p \oplus q} \leftarrow r_{p \oplus q} + \mbox{relatedness}(C_{p \oplus q}[i], C_q[j])$\;
												}
										}
									
										$\mathcal{S} \leftarrow \mathcal{S} \cup \{(C_{p \oplus q}, i_p, n_p + n_q - l, r_{p \oplus q})\}$\;
								}
						}
				}
		}
}

\For{$j \in \{1,2, ..., m\}$}
{
		$\mathcal{Q}_j \leftarrow \lbrace (C, i, d, r) \mid (C, i, d, r) \in \mathcal{S}, j \in \{i,i+1,...,i+d-1\} \rbrace$\;		
		
		$\mathcal{Q}_j \leftarrow$ the top $k$ configurations obtained by sorting the configurations in $\mathcal{Q}_j$ by their length (descending)\;
		
		$ps_{w_j} \leftarrow$ the predominant sense obtained by using a majority voting scheme on $\mathcal{Q}_j$\;
		
		$G[j] \leftarrow ps_{w_j}$\;
}

\BlankLine
\textbf{Output}: 

$G = (ps_{w_1}, ps_{w_2},..., ps_{w_m}), ps_{w_i} \in S_{w_i},$ $\forall i \in \{1,2,...,m\}$ -- the global configuration of senses returned by the algorithm.
}
\end{algorithm*}
We next present how we adapt the Shotgun sequencing technique for the task of global WSD. We will make a few observations along the way that will lead to a simplified method, namely \emph{ShotgunWSD}, which is formally presented in Algorithm~\ref{alg_ShotgunWSD}. We use the following notations in Algorithm~\ref{alg_ShotgunWSD}. An array (or an ordered set of elements) is denoted by $X = (x_1, x_2,....,x_m)$ and the length of $X$ is denoted by $|X| = m$. Arrays are considered to be indexed starting from position $1$, thus $X[i] = x_i, \forall i \in \{1,2,...m\}$. 

Our goal is to find a configuration of senses $G$ for the whole document $D$, that matches the ground-truth configuration produced by human annotators. A \emph{configuration of senses} is simply obtained by assigning a sense to each word in a text. In this work, the senses are selected from WordNet~\cite{Miller-WN-1995,Fellbaum-WN-1998}, according to steps $7$-$8$ of Algorithm~\ref{alg_ShotgunWSD}. Naturally, we will consider that the sense configuration of the whole document corresponds to the long DNA strand that needs to be sequenced. In this context, sense configurations of short context windows (less than $10$ words) will correspond to the short DNA reads. A crucial difference here is that we know the location of the context windows in the whole document from the very beginning, so our task will be much easier compared to Shotgun sequencing (we do not need to use a backbone solution for the alignment of short sense configurations). At every possible location in the text document (step $12$), we select a window of $n$ words. The window length $n$ is an external parameter of our algorithm that can be tuned for optimal results. For each context window we will compute all possible sense configurations (steps $14$-$15$). A score is assigned to each sense configuration by using the semantic relatedness between word senses (steps $16$-$19$), as described by \newcite{Patwardhan-CICLing-2003}. We alternatively employ two measures to compute the semantic relatedness, one is the extended Lesk measure~\cite{Banerjee-CICLING-2002,Banerjee-IJCAI-2003} and the other is a simple approach based on deriving sense embeddings from word embeddings~\cite{Mikolov-NIPS-2013}. Both methods are described in Section~\ref{sec_SemanticRelatedness}. We will keep the top scoring sense configurations (step $21$) for the assembly phase (steps $23$-$39$). In step $21$, we use an internal parameter $c$ in order to determine exactly how many sense configurations are kept per context window. Another important remark is that we assume that the BF algorithm used to obtain sense configurations for short windows does not produce output errors, so it is not necessary to use a distance measure in order to find overlaps for merging configurations. We simply check if the suffix of a former configuration coincides with the prefix of a latter configuration in order to join them together (steps $29$-$33$). The length $l$ of the suffix and the prefix that get overlapped needs to be greater then zero, so at least one sense choice needs to coincide. We gradually consider shorter and shorter suffix and prefix lengths starting with $l = min\{4,n-1\}$ (step $23$). Sense configurations are assembled in order to obtain longer configurations (step $34$), until none of the resulted configurations can be further merged together. When merging, the relatedness score of the resulting configuration needs to be recomputed (steps $36$-$38$), but we can take advantage of some of the previously computed scores (step $35$). Longer configurations indicate that there is an agreement (regarding the chosen senses) that spans across a longer piece of text. In other words, longer configurations are more likely to provide correct sense choices, since they inherently embed a higher degree of agreement among senses. After the configuration assembly phase, we start assigning the sense to each word in the document (step $40$). Based on the assumption that longer configurations provide better information, we build a ranked list of sense configurations for each word in the document (step $42$). Naturally, for a given word, we only consider the configurations that contain the respective word (step $41$). Finally, the sense of each word is given by a majority vote on the top $k$ configurations from its ranked list (steps $43$-$44$). The number of sense configurations $k$ is an external parameter of our approach, and it can be tuned for optimal results.
 
\subsection{Semantic Relatedness}
\label{sec_SemanticRelatedness} 

For a sense configuration of $n$ words, we compute the semantic relatedness between each pair of selected senses. We use two different approaches for computing the relatedness score and both of them are based on WordNet semantic relations. In this context, we essentially need to compute the semantic relatedness of two WordNet synsets. For each synset we build a disambiguation vocabulary by extracting words from the WordNet lexical knowledge base, as follows. Starting from the synset itself, we first include all the synonyms that form the respective synset along with the content words of the gloss (examples included). We also include into the disambiguation vocabulary words indicated by specific WordNet semantic relations that depend on the part-of-speech of the word. More precisely, we have considered hyponyms and meronyms for nouns. For adjectives, we have considered similar synsets, antonyms, attributes, pertainyms and related (see also) synsets. For verbs, we have considered troponyms, hypernyms, entailments and outcomes. Finally, for adverbs, we have considered antonyms, pertainyms and topics. These choices have been made because previous studies~\cite{Banerjee-IJCAI-2003,Hristea-2008} have come to the conclusion that using these specific relations for each part-of-speech seems to provide useful information in the WSD process. The disambiguation vocabulary generated by the WordNet feature selection described so far needs to be further processed in order to obtain the final vocabulary. 
The first processing step is to eliminate the stopwords. The remaining words are stemmed using the Porter stemmer algorithm~\cite{Porter-1980}. 
The resulted stems represent the final set of features that we use to compute the relatedness score between two synsets. The two measures that we employ for computing the relatedness score are described next. 

\subsubsection{Extended Lesk Measure}

The original Lesk algorithm~\cite{Lesk-1986} only considers one word overlaps among the glosses of a target word and those that surround it in a given context. \newcite{Banerjee-CICLING-2002} note that this is a significant limitation because dictionary glosses tend to be fairly short and they fail to provide sufficient information to make fine grained distinctions required for WSD. Therefore, \newcite{Banerjee-IJCAI-2003} introduce a measure that takes as input two WordNet synsets and returns a numeric value that quantifies their degree of semantic relatedness by taking into consideration the glosses of related WordNet synsets as well. Moreover, when comparing two glosses, the extended Lesk measure considers overlaps of multiple consecutive words, based on the assumption that the longer the phrase, the more representative it is for the relatedness of the two synsets. Given two input glosses, the longest overlap between them is detected and then replaced with a unique marker in each of the two glosses. The resulted glosses are then again checked for overlaps, and this process continues until there are no more overlaps. The lengths of the detected overlaps are squared and added together to obtain the score for the given pair of glosses. Depeding on the WordNet relations used for each part-of-speech, several pairs of glosses are compared and summed up together to obtain the final relatedness score. However, if the two words do not belong to the same part-of-speech, we only use their WordNet glosses and examples. Further details regarding this approach are provided by \newcite{Banerjee-IJCAI-2003}.

\subsubsection{Sense Embeddings} 

A simple approach based on word embeddings is employed to measure the semantic relatedness of two synsets. \emph{Word embeddings}~\cite{Bengio-JMLR-2003,Collobert-ICML-2008,Mikolov-NIPS-2013} represent each word as a low-dimensional real valued vector, such that related words reside in close vicinity in the generated space. We have used the pre-trained word embeddings computed by the \emph{word2vec} toolkit~\cite{Mikolov-NIPS-2013} on the Google News data set using the Skip-gram model. The pre-trained model contains $300$-dimensional vectors for $3$ million words and phrases.

The relatedness score between two synsets is computed as follows. For each word in the disambiguation vocabulary that represents a synset, we compute its word embedding vector. Thus, we obtain a cluster of word embedding vectors for each given synset. Sense embeddings are then obtained by computing the centroid of each cluster as the median of all the word embeddings in the respective cluster. We can naturally assume that some of the words in the cluster may actually be outliers. Thus, we believe that using the (geometric) median instead of the mean is more appropriate, as the mean is largely influenced by outliers. Finally, the semantic relatedness of two synsets is simply given by the cosine similarity between their cluster centroids.

It is important to note that an approach based on the mean of word vectors to construct sense embeddings is used by \newcite{Bhingardive-NAACL-2015}, but with a slightly different purpose than ours, namely to determine which synset better fits a target word, assuming that this synset should correspond to the most common sense of the respective word. As such, they completely disregard the context of the target word. Different from their approach, we are trying to find how related two synsets of distinct words that appear in the same context window are. Furthermore, the empirical results presented in Section~\ref{sec_Experiments} show that our approach yields better performance than the MCS estimation method of \newcite{Bhingardive-NAACL-2015}, thus putting a greater gap between the two methods.

\section{Experiments and Results}
\label{sec_Experiments}

\subsection{Data Sets}

We compare our global WSD algorithm with several state-of-the-art unsupervised WSD methods using the same test data as in the works presenting them. 

We first compare ShotgunWSD with two state-of-the-art approaches~\cite{Schwab-WET-2013,Chen-EMNLP-2014} and the MCS baseline, on the SemEval 2007 coarse-grained English all-words task~\cite{SemEval-task7-2007}. The SemEval 2007 coarse-grained English all-words data set\footnote{http://nlp.cs.swarthmore.edu/semeval/tasks/index.php} is composed of $5$ documents that contain $2269$ ambiguous words ($1108$ nouns, $591$ verbs, $362$ adjectives, $208$ adverbs) altogether. 
We also compare our approach with the MCS estimation method of \newcite{Bhingardive-NAACL-2015}, the MCS baseline and the extended Lesk algorithm~\cite{Torres-Lesk-2009} on the Senseval-2 English all-words~\cite{Senseval2-2001} and the Senseval-3 English all-words~\cite{Senseval3-2004} data sets. The Senseval-2 data set\footnote{\label{note1}http://web.eecs.umich.edu/$\sim$mihalcea/downloads.html} is composed of $3$ documents that contain $2473$ ambiguous words ($1136$ nouns, $581$ verbs, $457$ adjectives, $299$ adverbs), while the Senseval-3 data set\footnoteref{note1} is composed of $3$ documents that contain $2081$ ambiguous words ($951$ nouns, $751$ verbs, $364$ adjectives, $15$ adverbs).

\subsection{Parameter Tuning}

\begin{figure}
\begin{center}
\includegraphics[width=0.92\linewidth]{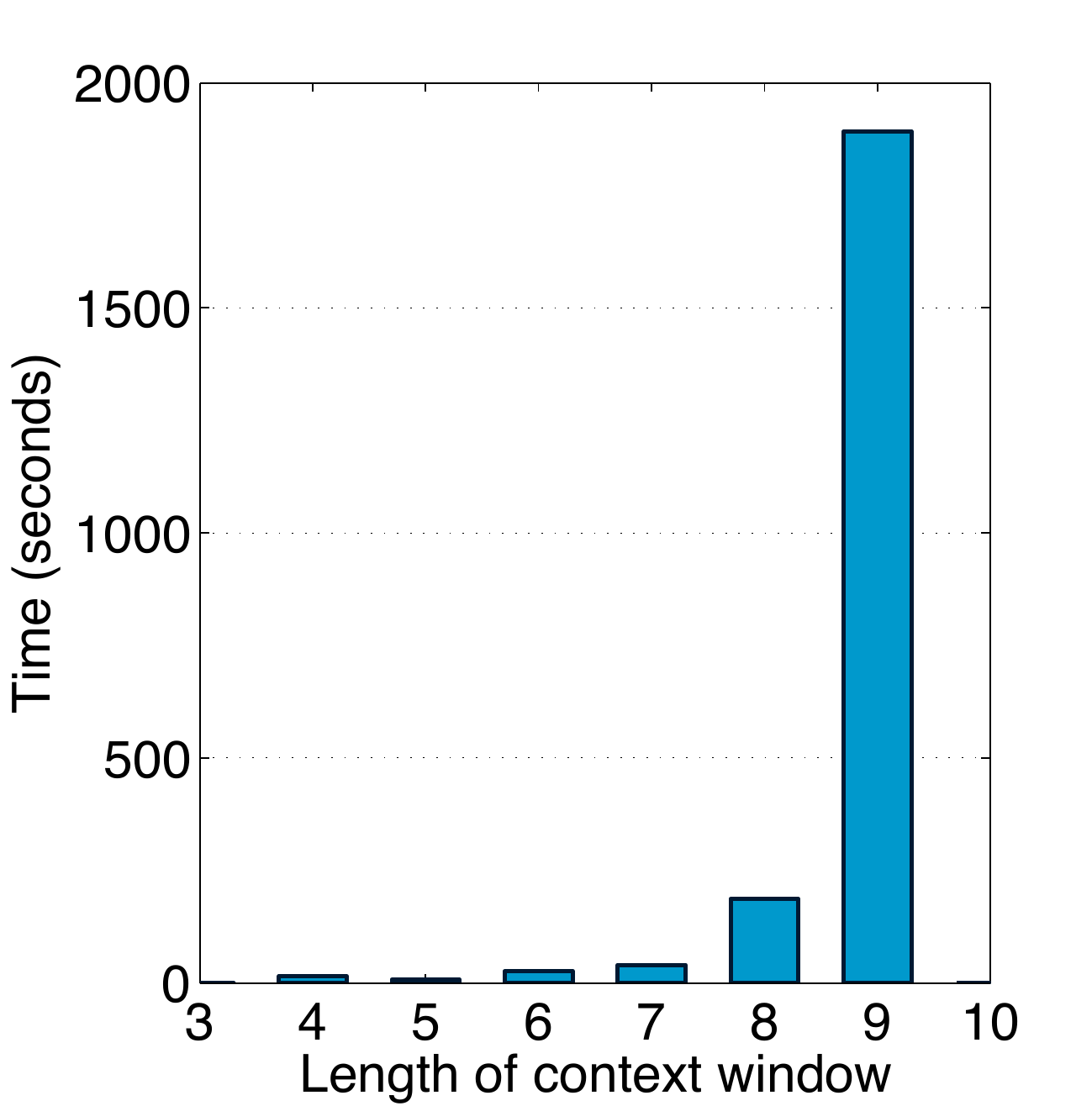}
\end{center}
\vspace*{-0.8em}
\caption{The running times (in seconds) of ShotgunWSD based on sense embeddings, on the first document of SemEval 2007, using various context window lengths $n \in \{4,5,6,7,8,9\}$. The reported times were measured on a computer with Intel Core i7 $3.4$ GHz processor and $16$ GB of RAM using a single Core.}
\label{fig_tuning_n}
\vspace*{-0.4em}
\end{figure}

As \newcite{Schwab-WET-2013}, we tune our parameters on the first document of SemEval 2007. We first set the value of the internal parameter $c$ to $20$ without specifically tuning it. Using this value for $c$ gives us a reasonable amount of configuration choices for the subsequent steps, without using too much space and time. For tuning the parameters $n$ and $k$, we employ sense embeddings for computing the semantic relatedness score. We begin by tuning the length of the context windows $n$. It is important to note that the upper bound accuracy of ShotgunWSD is given by the brute-force algorithm that explores every possible configuration of senses. Intuitively, we will get closer and closer to this upper bound as we use longer and longer context windows. However, the main decision factor is the time, which grows exponentially with respect to the length of the windows. Figure~\ref{fig_tuning_n} illustrates the time required by our algorithm to disambiguate the first document in SemEval 2007, for increasing window lengths in the range $4$-$9$. The algorithm runs in about $15$ seconds for $n = 4$ and in about $1892$ seconds for $n = 9$, so it becomes nearly $120$ times slower from using context windows of length $4$ to context windows of length $9$. As the algorithm runs in a reasonable amount of time for $n = 8$ ($187$ seconds), we choose to use context windows of $8$ words throughout the rest of the experiments.

\begin{figure}
\begin{center}
\includegraphics[width=1.0\linewidth]{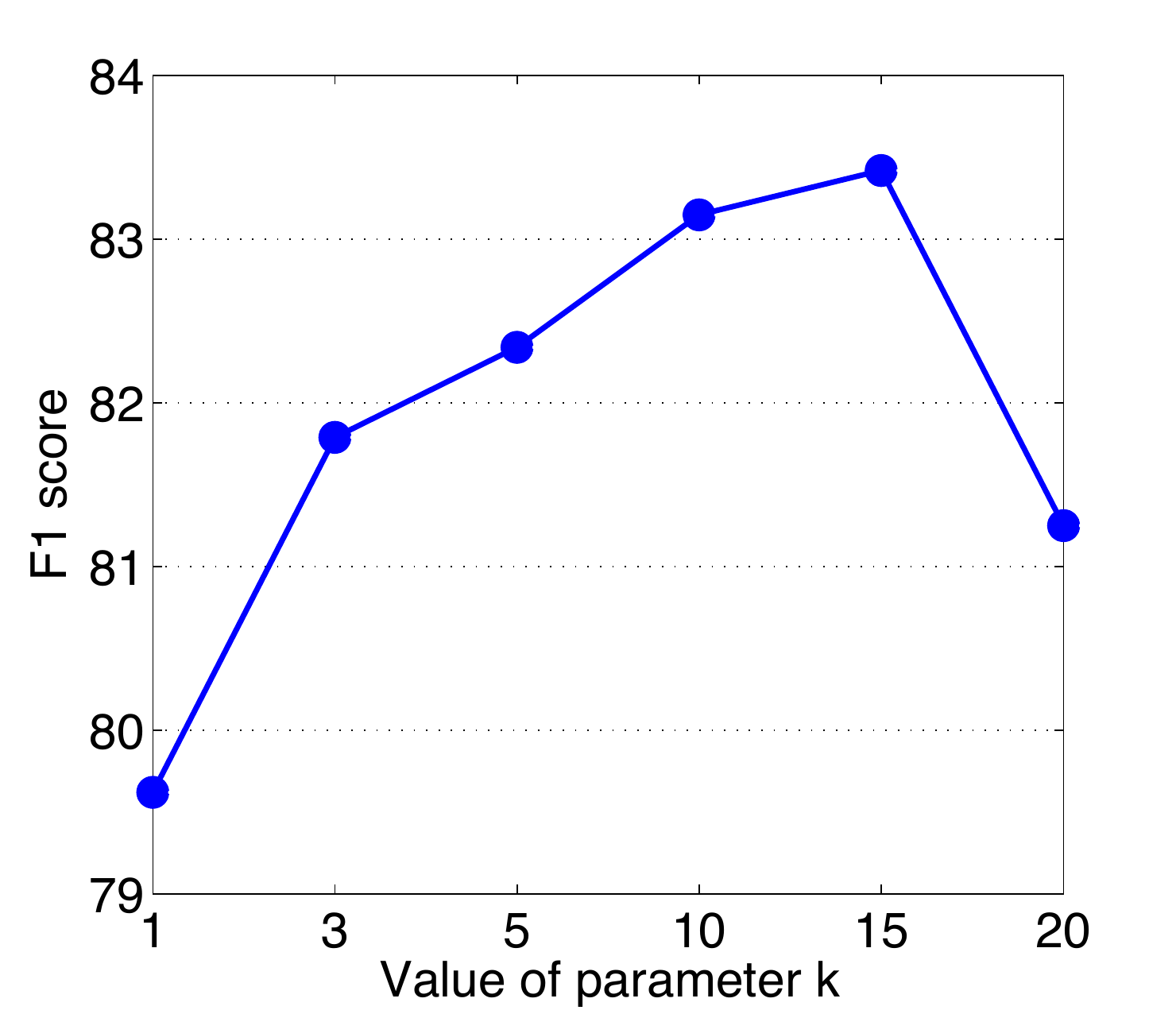}
\end{center}
\vspace*{-0.8em}
\caption{The $F_1$ scores of ShotgunWSD based on sense embeddings on the first document of SemEval 2007, using different values for the parameter $k \in \{1,3,5,10,15,20 \}$.}
\vspace*{-0.4em}
\label{fig_tuning_k}
\end{figure}

The parameter $k$ has almost no influence on the running time of the algorithm, so we tune this parameter with respect to the $F_1$ score obtained on the first document of SemEval 2007. We try out several values of $k$ in the set $\{1,3,5,10,15,20 \}$ and the results are shown in Figure~\ref{fig_tuning_k}. The best $F_1$ score ($83.42\%$) is obtained for $k = 15$. Hence, we choose to assign the final sense for each word using a majority vote based on the top $15$ configurations. 

To summarize, all the results of ShotgunWSD on SemEval 2007, Senseval-2 and Senseval-3 are reported using $n = 8$ and $k = 15$. We hereby note that better results in terms of accuracy can probably be obtained by trying out other values for these parameters on each data set. However, tuning the parameters on a single document from SemEval 2007 ensures that we avoid overfitting to a particular data set.

\subsection{Results on SemEval 2007}

\begin{table}[h]
\small{
\begin{center}
\begin{tabular}{|l|r|}
\hline
Method 																										& $F_1$ Score\\
\hline
Most Common Sense																					& $78.89\%$\\
Genetic Algorithms~\cite{Schwab-WET-2013} 											& $74.53\%$\\
Simulated Annealing~\cite{Schwab-WET-2013} 											& $75.18\%$\\
Ant Colony~\cite{Schwab-WET-2013} 														& $79.03\%$\\
S2C Unsupervised~\cite{Chen-EMNLP-2014}												& $75.80\%$\\
\hline
ShotgunWSD + Extended Lesk																		& $79.15\%$\\
ShotgunWSD + Sense Embeddings																& $79.68\%$\\
\hline
\end{tabular}
\end{center}
\caption{\label{tab_SemEval2007_Results} The $F_1$ scores of various unsupervised state-of-the-art WSD approaches, compared to the $F_1$ scores of ShotgunWSD based on the extended Lesk measue and ShotgunWSD based on sense embeddings, on the SemEval 2007 coarse-grained English all-words task. The results reported for both ShotgunWSD variants are obtained for windows of $n=8$ words and a majority vote on the top $k=15$ configurations.}
}
\end{table}

We first conduct a comparative study on the SemEval 2007 coarse-grained English all-words task in order to evaluate our ShotgunWSD algorithm. As described in Section~\ref{sec_SemanticRelatedness}, we use two different approaches for computing the semantic relatedness scores, namely extended Lesk and sense embeddings. We compare our two variants of ShotgunWSD with several algorithms presented in~\cite{Schwab-COLING-2012,Schwab-WET-2013}, namely a Genetic Algorithm, Simulated Annealing, and Ant Colony Optimization. We also include in the comparison an approach based on sense embeddings~\cite{Chen-EMNLP-2014}. All the approaches comprised in the evaluation are unsupervised. We compare them with the MCS baseline which is based on human annotations. The $F_1$ scores on SemEval 2007 are presented in Table~\ref{tab_SemEval2007_Results}. Among the state-of-the-art methods, it seems that the Ant Colony Optimization algorithm, based on a weighted voting scheme~\cite{Schwab-WET-2013}, is the only method able to surpass the MCS baseline. The unsupervised S2C approach gives lower results than the MCS baseline, but \newcite{Chen-EMNLP-2014} report better results in a semi-supervised setting. Both variants of ShotgunWSD yield better results than the MCS baseline ($78.89\%$) and the Ant Colony Optimization algorithm ($79.03\%$). Indeed, we obtain an $F_1$ score of $79.15\%$ when using the extended Lesk measure and an $F_1$ score of $79.68\%$ when using sense embeddings. We can also point out that ShotgunWSD gives slightly better results when sense embeddings are used instead of the extended Lesk method.

An important remark is that we have tuned the parameter $k$ on the first document included in the test set, following the same evaluation procedure as \newcite{Schwab-WET-2013}. Although this brings us to a fair comparison with \newcite{Schwab-WET-2013}, it might also raise suspicions of overfitting the parameter $k$ to the test set. Hence, we have tested all values of $k$ in $\{1,3,5,10,15,20 \}$ for ShotgunWSD based on word embeddings, and we have always obtained results above $79\%$, with the top score of $79.77\%$ for $k=10$. 

\subsection{Results on Senseval-2}

\begin{table}[h]
\small{
\begin{center}
\begin{tabular}{|l|r|}
\hline
Method 																										& $F_1$ Score\\
\hline
Most Common Sense																					& $60.10\%$\\
MCS Estimation~\cite{Bhingardive-NAACL-2015} 										& $52.34\%$\\
Extended Lesk~\cite{Torres-Lesk-2009}														& $54.60\%$\\
\hline
ShotgunWSD + Extended Lesk																		& $55.78\%$\\
ShotgunWSD + Sense Embeddings																& $57.55\%$\\
\hline
\end{tabular}
\end{center}
\caption{\label{tab_Senseval2_Results} The $F_1$ scores of an unsupervised WSD approach and the extended Lesk mesure, compared to the $F_1$ scores of ShotgunWSD based on the extended Lesk measue and ShotgunWSD based on sense embeddings, on the Senseval-2 English all-words data set. The results reported for both ShotgunWSD approaches are obtained for windows of $n=8$ words and a majority vote on the top $k=15$ configurations.}
}
\end{table}

We compare the two alternative forms of ShotgunWSD with the MCS baseline, the MCS estimation method of \newcite{Bhingardive-NAACL-2015} and the extended Lesk measure~\cite{Torres-Lesk-2009} on the Senseval-2 English all-words data set. As shown in Table~\ref{tab_Senseval2_Results}, the ShotgunWSD based on sense embeddings obtains an $F_1$ score that is almost $5\%$ better than the $F_1$ score of \newcite{Bhingardive-NAACL-2015}, while the ShotgunWSD based on extended Lesk gives an $F_1$ score that is around $1\%$ better than the $F_1$ score reported by \newcite{Torres-Lesk-2009}. It is important to note that \newcite{Torres-Lesk-2009} apply the extended Lesk measure by performing the brute-force search at the sentence level, hence it is not surprising that we are able obtain better results. However, our best ShotgunWSD approach ($57.55\%$) is still under the MCS baseline ($60.10\%$).

\subsection{Results on Senseval-3}

\begin{table}[h]
\small{
\begin{center}
\begin{tabular}{|l|r|}
\hline
Method 																										& $F_1$ Score\\
\hline
Most Common Sense																					& $62.30\%$\\
MCS Estimation~\cite{Bhingardive-NAACL-2015} 										& $43.28\%$\\
Extended Lesk~\cite{Torres-Lesk-2009}														& $49.60\%$\\
\hline
ShotgunWSD + Extended Lesk																		& $57.89\%$\\
ShotgunWSD + Sense Embeddings																& $59.82\%$\\
\hline
\end{tabular}
\end{center}
\caption{\label{tab_Senseval3_Results} The $F_1$ scores of an unsupervised WSD approach and the extended Lesk mesure, compared to the $F_1$ scores of ShotgunWSD based on the extended Lesk measue and ShotgunWSD based on sense embeddings, on the Senseval-3 English all-words data set. The results reported for both ShotgunWSD approaches are obtained for windows of $n=8$ words and a majority vote on the top $k=15$ configurations.}
}
\end{table}

We also compare the two variants of ShotgunWSD with the MCS baseline, the MCS estimation method of \newcite{Bhingardive-NAACL-2015} and the extended Lesk measure~\cite{Torres-Lesk-2009} on the Senseval-3 English all-words data set. The $F_1$ scores are presented in Table~\ref{tab_Senseval3_Results}. The empirical results show that both ShotgunWSD variants give considerably better results compared to the MCS estimation method of \newcite{Bhingardive-NAACL-2015}. By using sense embeddings in a completely different way than \newcite{Bhingardive-NAACL-2015}, we are able to report an $F_1$ score of $59.82\%$, which is much closer to the MCS baseline ($62.30\%$). With an $F_1$ score of $57.89\%$, the ShotgunWSD based on the extend Lesk measure brings an improvement of $8\%$ over the extended Lesk algorithm applied at the sentence level~\cite{Torres-Lesk-2009}.

\subsection{Discussion}

Considering all the experiments, we can conclude that ShotgunWSD gives better results (around $1\%$) when sense embeddings are used instead of the extended Lesk method. On one of the data sets, ShotgunWSD yields better performance than the MCS baseline. It is important to underline that the strong MCS baseline cannot be used in practice, since human input is required to indicate which sense of a word is the most frequent in a given text (a word's dominant sense will vary across domains and text genres). Corpora used for the evaluation of WSD methods usually contain this kind of annotations, but the MCS baseline will not work outside the annotated data. Therefore, we consider important even slightly outperforming the MCS baseline. Overall, our algorithm compares favorably to other state-of-the-art unsupervised WSD methods~\cite{Schwab-WET-2013,Chen-EMNLP-2014,Bhingardive-NAACL-2015} and to the extended Lesk measure~\cite{Banerjee-CICLING-2002,Torres-Lesk-2009}.

Regarding the performance of our algorithm, an interesting question that arises is how much does the assembly phase help. We look to investigate this further in future work, but we can carry out a small experiment to provide a quick answer to this question. We consider the ShotgunWSD variant based on sense embeddings without changing its parameters, and we remove the assembly phase completely. Therefore, the algorithm will no longer produce configurations of length greater than $8$, as the parameter $n$ is set to $8$. We have evaluated this stub algorithm on SemEval 2007 and we have obtained a lower $F_1$ score ($77.61\%$). This indicates that the assembly phase in Algorithm~\ref{alg_ShotgunWSD} boosts the performance by nearly $2\%$. More experiments are required to make sure that the performance boost is consistent across data sets.

\section{Conclusions and Future Work}
\label{sec_Conclusions}

In this paper, we have introduced a novel unsupervised global WSD algorithm inspired by the Shotgun genome sequencing technique~\cite{Shotgun-1981}. Compared to other bio-inspired WSD methods~\cite{Schwab-COLING-2012,Schwab-WET-2013}, our algorithm has only two parameters. Furthermore, our algorithm is deterministic, obtaining the same result for a given set of parameters and input document. The empirical results indicate that our algorithm can obtain better performance than other state-of-the-art unsupervised WSD methods~\cite{Schwab-WET-2013,Chen-EMNLP-2014,Bhingardive-NAACL-2015}. Although the fact that ShotgunWSD is deterministic brings several advantages, it is also a key difference from our source of inspiration, Shotgun sequencing, which is a non-deterministic technique.

In future work, we aim to investigate if training sense embeddings instead of deriving them from pre-trained word embeddings could yield better accuracy. Another promising direction is to compute the semantic relatedness of sense configurations based on the sum of sense tuples instead of sense pairs. An approach to combine the two semantic relatedness approaches independently used by ShotgunWSD, namely the extended Lesk measure and sense embeddings, is also worth exploring in the future.


\section*{Acknowledgments}
We thank the reviewers for their helpful suggestions. We also thank Alexandru I. Tomescu from the University of Helsinki for insightful comments about the Shotgun sequencing technique.

\bibliography{references}
\bibliographystyle{eacl2017}

\end{document}